\documentclass[10pt,twocolumn,letterpaper]{article}

\usepackage{wacv}

\usepackage{booktabs}
\usepackage{amsmath}
\usepackage{graphicx}
\graphicspath{{figs/}}

\definecolor{wacvblue}{rgb}{0.21,0.49,0.74}
\usepackage[pagebackref,breaklinks,colorlinks,allcolors=wacvblue]{hyperref}
\usepackage[capitalize]{cleveref}

\def\wacvPaperID{40}
\def\confName{WACV}
\def\confYear{2027}

\title{UniQuery: Unifying Vision Models with a Joint Query Head\\for Cross-Domain Pedestrian Attribute Recognition}
\author{Nguyen Dang Duy Tan$^{1}$\quad Linh T. T. Tran$^{2}$\quad Huynh Nguyen Phat$^{3}$\quad Vo Thanh Nguyen$^{4}$\\
$^{1}$AI VIET NAM\qquad $^{2}$University of Information Technology, VNU-HCM\\
$^{3}$University of Science, VNU-HCM\\
$^{4}$Ho Chi Minh City University of Technology and Engineering\\
{\tt\small nddtan2011@gmail.com}}

\begin{document}
\maketitle

\begin{abstract}
Pedestrian Attribute Recognition (PAR) predicts semantic attributes of individuals, such as gender, age, clothing and accessories, and its descriptions support attribute-based person retrieval and person re-identification. However, accurately recognizing fine-grained attributes across domains remains challenging due to occlusion, appearance variations, and shifts in image distributions. Ensembles of independently trained foundation encoders combine their logits with fixed fusion weights, so every attribute uses the same mix and the encoders are never trained together. We propose UniQuery, which instead jointly fine-tunes DINOv3~\cite{dinov3} and CLIP~\cite{clip} for attribute-specific fusion. It combines two query heads, one for each encoder, with a joint query head whose attribute queries attend to tokens from both encoders, and the prediction is the average of the logits of the three query heads. We evaluate UniQuery on the benchmark of the UPAR Challenge 2027 at the WACV 2027 Real-World Surveillance workshop, where models are trained on PA100k~\cite{pa100k}, PETA~\cite{peta} and Market-1501~\cite{market} with a shared set of attributes. In leave-one-dataset-out experiments, UniQuery outperforms the independent ensemble in Challenge\_Avg on both held-out datasets. Ablations show that learning fusion over independently trained encoders does not improve upon logit averaging in the mean over the two held-out datasets, while freezing the encoders lowers the Challenge\_Avg of UniQuery. On the hidden test set of the benchmark, UniQuery achieves a Challenge\_Avg of 0.8001.
\end{abstract}

\begin{figure*}[t]
\centering
\includegraphics[width=\linewidth]{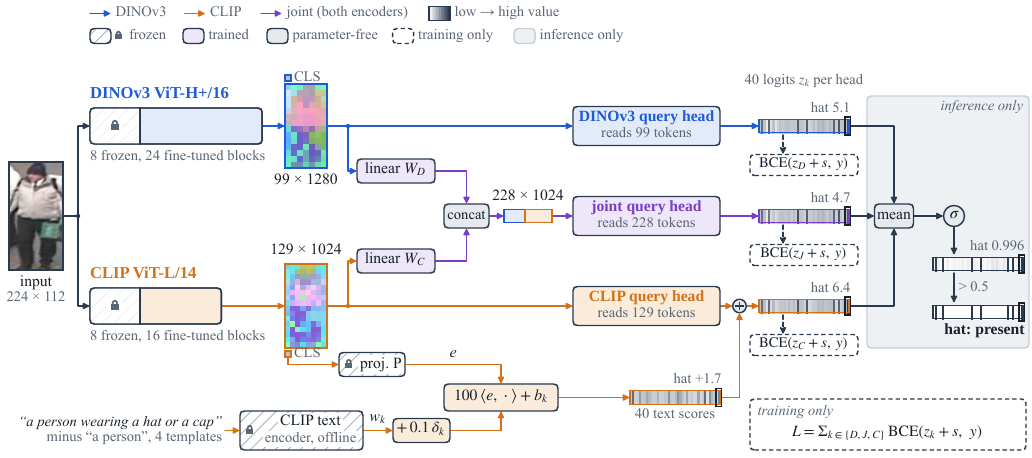}
\caption{UniQuery on one PETA image. Striped parts with a lock are frozen, light tints mark trained parts, gray boxes are operations without parameters, and token cells are colored by the PCA of the patch tokens. In the text-score branch, $P$ is the pretrained CLIP projection, $e$ the projected and normalized CLS token, $w_k$ the vector of the sentence for attribute $k$ minus that of \texttt{a person}, and $\delta_k$ and $b_k$ are learned. In training, each query head has its own loss with the shift $s$ (\cref{eq:loss}). At inference, the logits of the three query heads are averaged and compared with a threshold of 0.5. The final model, trained on the training and validation splits, saw this image in training.}
\label{fig:arch}
\end{figure*}

\section{Introduction}
\label{sec:intro}

Pedestrian analysis plays an important role in intelligent video surveillance and online retail applications. A key task, Pedestrian Attribute Recognition (PAR), predicts semantic attributes of individuals from images, providing interpretable descriptions that support attribute-based person retrieval and person re-identification. Despite recent progress, robust recognition remains challenging due to occlusion, low image resolution, label imbalance, and variations in appearance and imaging conditions~\cite{rap,upar,labelbalanced}. Recognition becomes harder still on a camera that was never seen in training, because resolution, viewpoint and lighting change, and each training dataset labels attributes by its own rules~\cite{upar}.

Foundation encoders pretrained on large and varied image collections~\cite{dinov3,clip} are a natural starting point for this shift. They are usually adapted to downstream tasks by fine-tuning one encoder with a small head, or by averaging the predictions of several independently trained models~\cite{deepens}. Earlier UPAR entries include a Swin-B encoder with a three-branch head that uses graph convolution and Transformer modules~\cite{upar23}, and an EfficientNetV2-L with one threshold tuned for each attribute on validation data~\cite{upar24}. C$^2$T-Net instead trains two encoders together and merges them into one vector shared by all attributes~\cite{c2tnet}. An average of independently fine-tuned encoders uses the same mix for every attribute, and its encoders are never trained together.

In this paper, we propose UniQuery (\cref{fig:arch}), which jointly fine-tunes DINOv3 and CLIP and gives each encoder its own query head. Alongside these two query heads, a joint query head combines features from both encoders through one query per attribute, so that each attribute can draw different information from the two encoders. At inference, we average the logits of the three query heads. To evaluate rigorously how effective the method is, we build an evaluation protocol on the UPAR Challenge 2027 benchmark, in which every design choice is tested on a dataset held out from training. The main contributions of this paper are below.
\begin{itemize}
\item We propose UniQuery, which fine-tunes DINOv3 and CLIP together and lets one query per attribute decide how to mix the two encoders. It outperforms an ensemble of independently fine-tuned encoders trained with the same loss on both held-out datasets, and it is also above the ensemble on the hidden test set, by a small margin. The gain needs the joint query head, because when it is removed at inference, the average of the other two query heads does not beat the ensemble in the two-fold mean (\cref{tab:main,tab:test}).
\item We show that, in our experiments, the mix needs to be learned together with fine-tuning the encoders. Fusion heads trained on top of independently fine-tuned encoders do not beat a plain logit average in the mean over the two held-out datasets, and with frozen encoders UniQuery falls below a single fine-tuned DINOv3 (\cref{tab:frozen,tab:swap}).
\item We analyze how the joint query head mixes the encoders. Each of its attention heads reads almost only the tokens of one encoder, and the number of attention heads given to DINOv3 or CLIP changes with the attribute. For most attributes, the joint query head is more accurate than the average of the two single-encoder query heads, and for attributes other than colors it often matches or beats the better one (\cref{fig:pick,fig:routing}).
\end{itemize}

\section{Related work}
\label{sec:related}

Cross-domain and unified learning have attracted researchers because high accuracy on an individual PAR dataset does not guarantee reliable performance in a new environment~\cite{upar,rethinkpar}. At the representation level, ASRL~\cite{asrl} reduces the classifier confusion caused by attribute correlations, by using one feature per attribute and a fixed orthogonal classifier. AttriVision~\cite{attrivision} fine-tunes both CLIP encoders so that an image embedding is close to the embeddings of short sentences describing its attribute values, and predicts each attribute by cosine similarity. Like AttriVision, UniQuery uses CLIP text embeddings, but it keeps the text encoder frozen. It only adds a text score to the logits of the CLIP query head, and this score comes from the similarity of the image to attribute sentences.

A related direction uses images of the target domain. AdaGPAR~\cite{adagpar} keeps its weights fixed at test time and classifies new images by their similarity to prototypes built from reliable target predictions. LDCD-PAR~\cite{ldcd} also trains on unlabeled target images, with an adversarial domain discriminator on local features. In contrast, UniQuery is a domain generalization method and uses no target-domain image.

Unified learning offers another route to transferable representations. UniHCP~\cite{unihcp} trains one ViT encoder and one decoder for many human-centric tasks, where each task has its own queries and PAR has one query per attribute. The PATH method of HumanBench~\cite{humanbench} pretrains one shared encoder on many human-centric datasets with one projector per task. CLEAR~\cite{clear} reuses its attribute recognition network for person retrieval by passing a sentence that describes the query attributes through a GPT language model and training only small adapters. Like UniHCP and query-based multi-label classifiers~\cite{q2l,mldecoder}, UniQuery uses queries that read the image tokens through cross-attention, but the queries of its joint query head read the tokens of two different encoders.

The two encoders of UniQuery differ in how they are pretrained. DINOv3~\cite{dinov3} learns by self-supervised learning on unlabeled images and gives high-quality patch features, while CLIP~\cite{clip} and its successors~\cite{metaclip,dfn,evaclip,siglip2,pe} learn to match images with captions on web data. The variety of this web data is the main source of the robustness of CLIP to distribution shift~\cite{clipdata}. Still, CLIP treats some clearly different images as similar, while a self-supervised encoder separates them~\cite{eyeswide}. Adding self-supervised features next to CLIP improves multimodal language models on vision-centric benchmarks~\cite{eyeswide,cambrian}, so we pair DINOv3 with CLIP.

A simple way to combine two encoders is an ensemble. Each encoder is fine-tuned alone into its own model, and the predictions of the two models are averaged at inference~\cite{deepens}. An ensemble only helps on images where the two models predict differently~\cite{krogh}. It also uses the same mix for every attribute. Weight averaging~\cite{soups,wiseft} cannot merge DINOv3 with CLIP, because it needs models with the same starting point. Distillation merges several encoders into one~\cite{amradio}, but the student is still worse than its teachers on some tasks, such as zero-shot classification against its image--text teacher~\cite{amradio,cradio4}. Stacking learns the mix on models that are already trained~\cite{stacking}. In UPAR 2024~\cite{upar24}, C$^2$T-Net~\cite{c2tnet} trained a Swin-B and an EVA ViT-L together and merged them into one vector for all attributes. UniQuery also trains the two encoders together, but each attribute has its own query that sets the mix, and each encoder keeps its own query head.

Attributes also differ in frequency, and most of the 40 attributes have few positive samples. A model trained in the usual way is biased toward predicting such attributes as absent~\cite{logitadj}. This bias lowers the mean balanced accuracy (mA), which averages over attributes the accuracy on positives and the accuracy on negatives with equal weight. It therefore also lowers Challenge\_Avg, the challenge metric, which is the harmonic mean of mA and instance F1. DeepMAR~\cite{deepmar} corrects for this by weighting the loss of each attribute by its positive ratio, and UniQuery also uses these weights. In UPAR 2024, one entry also tuned one threshold per attribute on validation images of the source domains~\cite{upar24}, so these thresholds may not hold on new cameras. Logit adjustment~\cite{logitadj} adds a shift based on the attribute frequency to the logits inside the training loss, so UniQuery keeps one threshold of 0.5 for all attributes at inference.

\section{The proposed method}
\label{sec:method}

\begin{figure}[t]
\centering
\includegraphics[width=\linewidth]{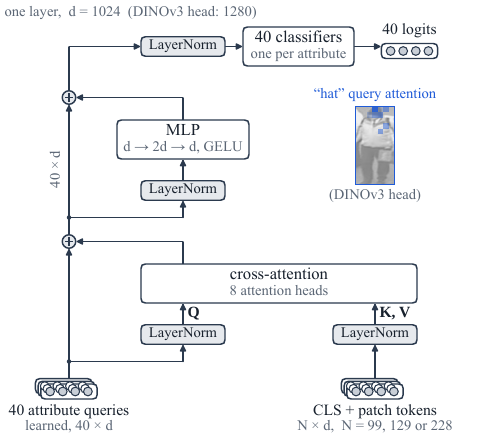}
\caption{One query head. The three query heads of UniQuery share this design and differ only in the tokens they read. The inset is the attention of the hat query in the DINOv3 query head on the image of \cref{fig:arch}, averaged over the eight attention heads.}
\label{fig:qhead}
\end{figure}

\Cref{fig:arch} shows the architecture of UniQuery. A cropped pedestrian image of $224\times112$ pixels passes through two encoders, DINOv3 ViT-H+/16~\cite{dinov3} and the OpenAI CLIP ViT-L/14 image encoder~\cite{clip}, which are fine-tuned together within one architecture. Three query heads read the tokens of the two encoders in three ways. The DINOv3 and CLIP query heads read only the tokens of their own encoder, and the joint query head reads the tokens of both. Each query head gives 40 logits, one per binary attribute. During training, each query head has its own loss, and at inference the logits of the three query heads are averaged.

\begin{figure}[t]
\centering
\includegraphics[width=0.68\linewidth]{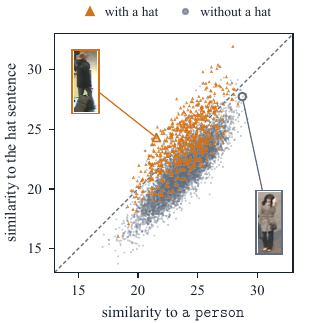}
\caption{The idea of the text score for the hat attribute, on PETA validation images. The text score of an image is the vertical distance from its point to the diagonal. The image without a hat is more similar to the hat sentence than the image with a hat, but it lies below the diagonal and so gets a lower text score.}
\label{fig:zsidea}
\end{figure}

The CLIP query head also uses the image--text matching that CLIP learned in pretraining. The text score of an attribute is the similarity of the image to a sentence that describes the attribute minus its similarity to the neutral sentence \texttt{a person} (\cref{fig:zsidea}). This score is added to the logit that the CLIP query head computes from the image tokens. The subtraction aims to remove the part that every attribute sentence shares, and the addition aims to bring the knowledge of CLIP about the attribute into this logit.

The tokens of DINOv3 and CLIP have different widths, so before the joint query head each kind of token passes through its own linear layer to reach a common width. The tokens of the two encoders are then concatenated into one sequence that the joint query head reads. Inside, each query head consists of a cross-attention layer, a multilayer perceptron (MLP) and a classifier~\cite{q2l,mldecoder} (\cref{fig:qhead}).

During training, each query head has its own loss. Let $D$, $J$ and $C$ denote the DINOv3, joint and CLIP query heads, and let $z_{k,a}$ be the logit of query head $k$ for attribute $a$. The label $y_a$ is smoothed with a factor of 0.05, and the mask $m_a$ removes the attributes that are not annotated in the dataset of the input image. Let $p_a$ be the frequency of attribute $a$ in the training data. The DeepMAR weight $w_a(y_a)$~\cite{deepmar} is $e^{1-p_a}$ for positives and $e^{p_a}$ for negatives. The shift $s_a=\tau\log\frac{p_a}{1-p_a}$ applies logit adjustment~\cite{logitadj} and is used only in training. The loss is
\begin{equation}
L=\sum_{k\in\{D,J,C\}}\;\sum_{a=1}^{40} m_a\,w_a(y_a)\,\mathrm{BCE}\big(z_{k,a}+s_a,\;y_a\big),
\label{eq:loss}
\end{equation}
with $\tau=0.32$. The DeepMAR weight moves the decision threshold by $1-2p_a$ on the logit scale, so the loss shifts the decision in two parts, $1-2p_a$ from $w_a$ and a part from $s_a$. We chose $\tau=0.32$ so that their sum fits, by least squares over the 40 attributes, the total shift of a post-hoc rule. In this rule, a model trained with the DeepMAR weights alone adds $0.42\,\big(\log\frac{1-p_a}{p_a}-(1-2p_a)\big)$ to its logits after training. The coefficient 0.42 of this rule was chosen on the leave-one-dataset-out runs.

During training, we freeze the patch embedding and the first eight blocks of each encoder and fine-tune the rest with AdamW for three epochs. We use a batch size of 64 and an exponential moving average (EMA) of the weights. The learning rate is $5\times10^{-5}$ at the top block, decays by 0.75 per block below, and is ten times higher in the query heads. The final model, which we evaluate on the hidden test set, is trained on the training and validation splits of all three datasets. At inference, the average logit of the three query heads passes through a sigmoid, and an attribute is predicted as present when the probability exceeds 0.5. Like an ensemble of the two separately trained encoders, UniQuery needs only two encoder forward passes per image. Its measured latency is about 3.0\,ms per image on an NVIDIA RTX 5090 GPU with batches of 128 images.

\section{Experiments}
\label{sec:exp}

We evaluated independently trained models, each query head of UniQuery, and UniQuery with its three query heads averaged. The average of the three query heads reaches a higher mean Challenge\_Avg than each query head alone and than the independent ensemble, so we use it as the final model.

\paragraph{Protocol.} To mimic the unseen cameras of the hidden test set, we use a leave-one-dataset-out protocol. In each fold, one dataset is held out completely, and the model sees none of its images during training. The model is trained on the training splits of the other two datasets and scored on the validation split of the held-out dataset. The PETA fold trains on PA100k and Market-1501 and is scored on the 6,963 PETA validation images. The Market fold trains on PA100k and PETA and is scored on the 16,458 Market-1501 validation images. We do not use the fold that holds out PA100k, because only 18.7k training images would remain. The metric is Challenge\_Avg, in percent. Fold models are trained like the final model, except that their training data exclude the validation splits. Each image passes through the model once, without flipping, as for the final model. Results are for one seed unless stated otherwise. The factor 0.42 used to set $\tau$ and the encoder paired with DINOv3 were chosen on these same folds, so the fold scores may be slightly optimistic for these choices. The hidden test set is the only evaluation that these choices do not affect.

\begin{table}[t]
\centering
\small
\setlength{\tabcolsep}{3pt}
\begin{tabular}{@{}lccc@{}}
\toprule
Model & PETA & Market & Avg. \\
\midrule
Independently trained DINOv3 & 81.46 & 86.31 & 83.89 \\
Independently trained CLIP & 81.44 & 85.50 & 83.47 \\
Ensemble of the two encoders, seed 1 & 82.45 & 86.30 & 84.38 \\
Ensemble of the two encoders, seed 2 & 82.29 & 86.33 & 84.31 \\
\midrule
UniQuery, DINOv3 query head only & 81.26 & 86.07 & 83.67 \\
UniQuery, CLIP query head only & 81.45 & 85.56 & 83.51 \\
UniQuery, joint query head only & 82.26 & 86.27 & 84.26 \\
UniQuery, DINOv3 + CLIP query heads & 82.24 & 86.32 & 84.28 \\
UniQuery, three query heads averaged & \textbf{82.54} & 86.39 & 84.46 \\
\midrule
Variant, seed 1 & 82.53 & \textbf{86.44} & \textbf{84.49} \\
Variant, seed 2 & 82.51 & 86.32 & 84.42 \\
Variant with two joint query heads & 82.51 & 86.43 & 84.47 \\
\bottomrule
\end{tabular}
\caption{Held-out Challenge\_Avg with the same loss as in \cref{eq:loss}. Avg.\ is the mean of the two folds, and the final setting is UniQuery with its three query heads averaged. The row ``DINOv3 + CLIP query heads'' averages these two query heads of the same model, without the joint query head. The variant adds a text score for clothing colors on the upper or lower body region, with a different set of sentences, and in training hides each encoder from the joint query head with probability 0.25.}
\label{tab:main}
\end{table}

\paragraph{UniQuery versus the independent ensemble.} \Cref{tab:main} compares UniQuery with an ensemble of the two encoders trained independently, with the same encoder architectures and the same loss. The logit average of the three query heads of UniQuery beats both seeds of the ensemble on both folds. This gain needs the joint query head. When the joint query head is removed at inference, the average of the DINOv3 and CLIP query heads of UniQuery is no longer above the ensemble in the two-fold mean. Used alone, the joint query head is above each single-encoder query head but it does not beat the ensemble either. The gain appears only when the joint query head is averaged with the two single-encoder query heads. On the hidden test set, both UniQuery models are also above the independent ensemble with the same loss (\cref{tab:test}).

\paragraph{Consistency.} UniQuery beats the ensemble by only 0.08 to 0.15 points in the mean, close to the 0.07-point gap between the two seeds of the ensemble. We therefore check whether the gain repeats in other training runs. The last three rows of \cref{tab:main} are two seeds of a UniQuery variant with a few small changes, given in the table caption, and one run of this variant with two joint query heads. They are not reruns of UniQuery with other seeds. All four runs, UniQuery and these three, have a two-fold mean above both seeds of the ensemble. Fold by fold, the four runs win all 8 comparisons with the two ensemble seeds on PETA and 7 of 8 on Market.

\begin{figure}[t]
\centering
\includegraphics[width=\linewidth]{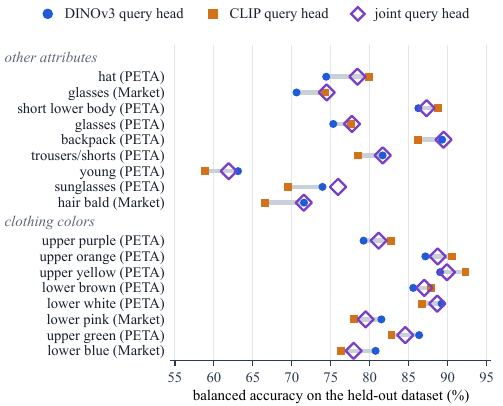}
\caption{The joint query head next to the two single-encoder query heads, for the attributes on which these two differ by at least 2 points and that have at least 50 positive images. Values are means over two seeds of the variant in \cref{tab:main}.}
\label{fig:pick}
\end{figure}

\begin{figure*}[t]
\centering
\includegraphics[width=\linewidth]{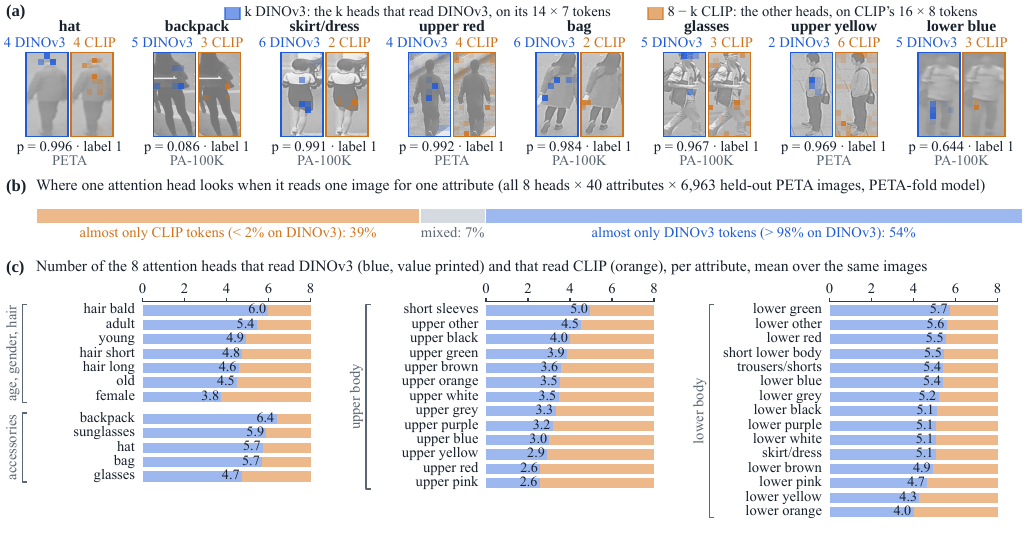}
\caption{How the joint query head uses the two encoders. (a) Attention of the attention heads that read DINOv3 and of those that read CLIP, on eight validation images that the final model saw in training. (b, c) Statistics over the 6,963 held-out PETA images, for the PETA-fold model.}
\label{fig:routing}
\end{figure*}

\paragraph{Attribute-specific mixing.} The joint query head does not mix the two encoders with one ratio for all attributes. \Cref{fig:pick} compares it with the two single-encoder query heads of the UniQuery variant in \cref{tab:main}. Over all attributes, both seeds and both folds, the joint query head is above the mean of the two single-encoder query heads in 82\% of cases. For attributes other than colors, such as age, gender, hair, clothing type and accessories, it matches or beats the better single-encoder query head in 69\% of cases, for example for backpack, sunglasses and trousers/shorts. For clothing colors, this happens in only 40\% of cases, and the joint query head often lies between the two. Its attention shows that it shares the work between the encoders by attribute (\cref{fig:routing}). On the held-out PETA images, in 93\% of the cases where an attention head reads an image for one attribute, more than 98\% of its attention falls on the tokens of a single encoder (\cref{fig:routing}b). We can therefore count, for each attribute, how many of the eight attention heads read DINOv3 (\cref{fig:routing}c). An attention head reads DINOv3 if more than half of its attention falls on DINOv3 tokens. Backpack uses 6.4 of the eight attention heads on DINOv3 on average, while red and pink upper clothing use only about 2.6. In general, lower-body and accessory attributes rely more on DINOv3, and upper-body colors rely more on CLIP.

\paragraph{Choice of DINOv3 and CLIP.} \Cref{tab:two} shows that the second encoder should differ from the first. A second fine-tuned copy of the same encoder adds at most 0.10 points in the mean, while a different encoder adds 0.4 to 0.6 points. Relative to DINOv3 alone, this gain comes from PETA. On Market, adding CLIP helps much less, and with $\tau=0.42$ the pair of DINOv3 and CLIP is even below DINOv3 alone. Our hypothesis is that two encoders pretrained in different ways are wrong on the same image less often than two fine-tuned copies of one encoder. To test this, we take the image--attribute pairs that one model gets wrong, measure the share that the other model also gets wrong, and average over both directions. Two fine-tuned copies of the same encoder trained with weighted BCE in \cref{tab:two} share 89\% to 93\% of their errors, while DINOv3 and CLIP share only 72\% on PETA and 77\% on Market-1501. To choose the encoder next to DINOv3, we tried 12 encoders on the PETA fold: CLIP, AIMv2~\cite{aimv2}, EVA02-CLIP~\cite{evaclip}, MetaCLIP2~\cite{metaclip2}, MobileCLIP2~\cite{mobileclip2}, DFN-CLIP~\cite{dfn}, two PE models~\cite{pe} and four SigLIP2 models~\cite{siglip2}. Each is fine-tuned alone and its logits are averaged with those of DINOv3. Every one of them shares only 66\% to 72\% of its errors with DINOv3, so the Challenge\_Avg of the pair follows that of the second encoder alone. Alone, CLIP is at least 0.5 points above every other encoder, and over five CLIP runs the pair of DINOv3 and CLIP reaches 81.04\% to 81.23\%. Only the pair with AIMv2-L falls in this range, and the other 10 encoders give lower pairs. The training recipe was tuned on DINOv3 and CLIP, so it may suit the other encoders less well. Inside UniQuery, replacing CLIP with SigLIP2-L/16 or C-RADIOv4-H~\cite{cradio4}, or replacing DINOv3 with C-RADIOv4-H, lowers the two-fold mean of Challenge\_Avg (\cref{tab:swap}). The drop comes mostly from PETA. On Market, the pair with SigLIP2-L is also lower, while the two pairs with C-RADIOv4-H are slightly above UniQuery.

\begin{table}[t]
\centering
\small
\setlength{\tabcolsep}{4pt}
\begin{tabular}{@{}lccc@{}}
\toprule
Encoders (logits averaged) & PETA & Market & Avg. \\
\midrule
\multicolumn{4}{@{}l}{\emph{Logit-adjusted loss, $\tau=0.42$}} \\
CLIP & 81.79 & 85.72 & 83.75 \\
CLIP + CLIP (two seeds) & 81.80 & 85.78 & 83.79 \\
DINOv3 & 81.14 & \textbf{86.78} & 83.96 \\
DINOv3 + CLIP & \textbf{82.29} & 86.49 & \textbf{84.39} \\
\midrule
\multicolumn{4}{@{}l}{\emph{Weighted BCE}} \\
CLIP & 80.68 & 85.01 & 82.84 \\
CLIP + CLIP (two seeds) & 80.69 & 85.12 & 82.90 \\
DINOv3 & 80.49 & 85.31 & 82.90 \\
DINOv3 + DINOv3 (two runs) & 80.66 & 85.35 & 83.00 \\
DINOv3 + CLIP & \textbf{81.08} & \textbf{85.54} & \textbf{83.31} \\
\bottomrule
\end{tabular}
\caption{Held-out Challenge\_Avg of separately fine-tuned encoders with averaged logits. These runs come from an earlier stage, trained with $\tau=0.42$ instead of 0.32 in \cref{eq:loss}, or with weighted BCE (\cref{eq:loss} without $s_a$), because only this stage has two fine-tuned copies of the same encoder. The two DINOv3 copies are two runs of the same recipe. In the weighted-BCE block, the CLIP and DINOv3 rows are the mean of the two copies.}
\label{tab:two}
\end{table}

\begin{table}[t]
\centering
\small
\setlength{\tabcolsep}{2pt}
\begin{tabular}{@{}lcccc@{}}
\toprule
Mix of the two models & PETA & Market & Avg. & Source \\
\midrule
Logit average, nothing learned & \textbf{82.45} & 86.30 & \textbf{84.38} & 89.46 \\
Learned ratio per attribute & 82.43 & \textbf{86.31} & 84.37 & 89.49 \\
\midrule
\multicolumn{5}{@{}l}{\emph{Logit average plus a correction learned from}} \\
\quad DINOv3 attribute vectors & 81.15 & 85.99 & 83.57 & 90.93 \\
\quad CLIP attribute vectors & 81.55 & 86.06 & 83.81 & 90.52 \\
\quad both, concatenated into an MLP & 81.27 & 85.92 & 83.60 & 91.02 \\
\quad both, ratio per attribute & 81.25 & 85.95 & 83.60 & 90.97 \\
\quad both, ratio per image & 81.01 & 85.86 & 83.44 & 91.14 \\
\bottomrule
\end{tabular}
\caption{Learning the mix on top of the two separately fine-tuned models of the ensemble, which stay fixed (held-out Challenge\_Avg). Corrections are computed by a linear layer and start at zero. They are trained with the loss of \cref{eq:loss} on 85\% of the validation images of the two source datasets (about 22k images in the PETA fold and 14k in the Market fold). The Source column is the two-fold mean Challenge\_Avg on the remaining 15\% of source images, which the fusion heads do not see in training.}
\label{tab:frozen}
\end{table}

\paragraph{A mix learned after training does not transfer.} The independent ensemble mixes its two models by averaging their logits, with the same equal ratio for every attribute. We test whether a better mix can be learned after the two models are trained, without joint training as in UniQuery. The two models stay fixed, and only a small fusion head on top of their outputs is trained (\cref{tab:frozen}). The first fusion head only learns one mixing ratio of the two logits per attribute. The other five add a correction to the logit average, computed from the attribute vectors of the query heads. An attribute vector is the vector that a query head produces for one attribute just before its classifier. In the two-fold mean, no fusion head beats the plain logit average, and the five with a correction lower Challenge\_Avg by 0.6 to 0.9 points. Their failure does not come from poor learning. On source images that they did not see in training, these five beat the logit average by 1.1 to 1.7 points. For these five fusion heads, a mix learned after training therefore fits the source datasets but does not carry over to a new dataset. This is why UniQuery learns its mix while it trains the two encoders.

\begin{table}[t]
\centering
\small
\setlength{\tabcolsep}{3.5pt}
\begin{tabular}{@{}llccc@{}}
\toprule
Encoder pair & Training & PETA & Market & Avg. \\
\midrule
DINOv3 + CLIP (UniQuery) & fine-tuned & \textbf{82.54} & 86.39 & \textbf{84.46} \\
DINOv3 + CLIP & frozen & 81.07 & 85.80 & 83.43 \\
\midrule
C-RADIOv4-H + CLIP & fine-tuned & 81.87 & 86.43 & 84.15 \\
DINOv3 + C-RADIOv4-H & fine-tuned & 82.11 & \textbf{86.50} & 84.31 \\
DINOv3 + SigLIP2-L & fine-tuned & 81.81 & 86.07 & 83.94 \\
\bottomrule
\end{tabular}
\caption{Freezing or replacing the encoders of UniQuery (held-out Challenge\_Avg). With frozen encoders, only 32.6M parameters are trained.}
\label{tab:swap}
\end{table}

\paragraph{The encoders need fine-tuning.} BRAVE~\cite{brave} keeps several pretrained encoders frozen and trains only a module whose learned queries read their tokens. We tried this with UniQuery by freezing both encoders and training only the three query heads and the two linear layers. UniQuery then loses 1.0 point in the mean and falls below independently fine-tuned DINOv3 (\cref{tab:swap}). The CLIP query head loses 2.7 points, while the DINOv3 query head loses only 0.6 points. In this task, the encoders therefore need fine-tuning, and freezing hurts CLIP more than DINOv3.

\begin{figure}[t]
\centering
\includegraphics[width=0.74\linewidth]{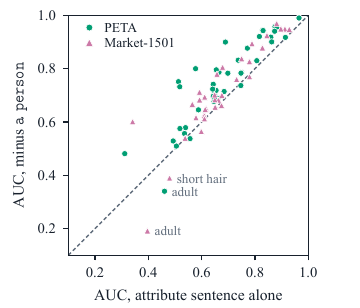}
\caption{Zero-shot AUC of the text score on validation images, one point per attribute. Points more than 0.05 below the diagonal are named.}
\label{fig:zsauc}
\end{figure}

\paragraph{The CLIP text score helps, but only slightly.} Every attribute sentence describes a person, so the similarity of an image to an attribute sentence shares a part with its similarity to \texttt{a person}. This shared part only says how much the image looks like a person, not whether the person has the attribute. Subtracting it should keep the part that is specific to the attribute. \Cref{fig:zsauc} tests this before training with the AUC (area under the ROC curve). The AUC is the probability that an image with the attribute gets a higher score than an image without it. On PETA, the subtraction raises the mean AUC from 0.68 to 0.76 and raises the AUC for 37 of the 40 attributes, and Market-1501 shows the same direction. After training, the benefit is much smaller. Removing the text score lowers Challenge\_Avg by only 0.15 points on PETA and 0.05 points on Market, about the size of the gap between two seeds. One possible reason is that once CLIP is fine-tuned on labeled data, the CLIP query head learns most of this information directly from the image tokens. The text score therefore carries correct attribute information but is a minor component, not the main source of the gain of UniQuery.

\begin{table}[t]
\centering
\small
\setlength{\tabcolsep}{3.5pt}
\begin{tabular}{@{}llc@{}}
\toprule
Model & Loss & Challenge\_Avg \\
\midrule
Independent ensemble & weighted BCE$^\dagger$ & 0.7893 \\
Independent ensemble & logit-adjusted & 0.7989 \\
UniQuery, variant & logit-adjusted & 0.7992 \\
UniQuery (final) & logit-adjusted & 0.8001 \\
\bottomrule
\end{tabular}
\caption{Challenge\_Avg on the hidden test set of the UPAR 2027 benchmark (scale 0 to 1). Final models are trained on the training and validation splits of all three datasets. All models use a threshold of 0.5 without post-processing, and the variant is the one in \cref{tab:main}. $^\dagger$With WiSE-FT~\cite{wiseft} for CLIP.}
\label{tab:test}
\end{table}

\section{Limitations and future work}
\label{sec:limits}

Our evaluation covers the PAR task of UPAR 2027, with its three datasets (PA100k, PETA and Market-1501) and its 40 attributes, and most of our results are based on a single training seed. A natural next step is to apply UniQuery to other PAR benchmarks and to other person-centric tasks, such as attribute-based person retrieval and person re-identification, where the attribute queries could be replaced by queries for each task. The joint query head is also not limited to two encoders. In principle, it can read the tokens of any group of encoders, so future work can study other pairs and larger groups, and test whether each attribute still relies mainly on one encoder. Repeating the main comparisons over more training seeds would also give a more precise estimate of the gain over the ensemble.

\section{Conclusion}
\label{sec:conclusion}

In this paper, we studied how to combine two foundation encoders for pedestrian attribute recognition on cameras that are not seen during training. We proposed UniQuery, which fine-tunes DINOv3 and CLIP within a single architecture and assigns a learned query to each attribute, so that each attribute can select the information it needs from the two encoders. The errors of DINOv3 and CLIP overlap less than those of two fine-tuned copies of the same encoder, which is consistent with their different pretraining, and combining them is more effective than combining two such copies. The main finding of this work is that the mixing of the two encoders should be learned jointly with their fine-tuning, through a joint query head placed alongside the query head of each encoder. When the joint query head is removed at inference, UniQuery no longer outperforms the independent ensemble in the mean over the two held-out datasets. Moreover, none of the fusion heads trained on top of independently trained models improves upon a plain logit average on a new dataset. Within the joint query head, nearly every attention head attends mainly to one encoder, and attributes of the lower body and accessories rely more on DINOv3, whereas upper-body colors rely more on CLIP. For cross-domain pedestrian attribute recognition, these results suggest that the assignment of encoders to attributes should be learned during training rather than fixed by a single mixing ratio shared by all attributes.

{\small
\bibliographystyle{ieeenat_fullname}
\bibliography{refs}
}

\end{document}